\documentclass{article} 
\usepackage[preprint]{colm2025_conference}

\usepackage{microtype}
\usepackage{hyperref}
\usepackage{url}
\usepackage{booktabs}

\usepackage{graphicx}
\usepackage{multirow}
\usepackage{tabularray}
\usepackage{subcaption} 
\usepackage{colortbl}
\usepackage[table]{xcolor}
\usepackage{amsmath}
\usepackage{algorithm}
\usepackage[noend]{algpseudocode}
\usepackage{amsthm}
\usepackage{amssymb}
\usepackage{caption}
\usepackage{wrapfig}

\usepackage{lineno}

\definecolor{darkblue}{rgb}{0, 0, 0.5}
\hypersetup{colorlinks=true, citecolor=darkblue, linkcolor=darkblue, urlcolor=darkblue}

\title{
Auto-Prompt Ensemble for LLM Judge
}

\author{
Jiajie Li\thanks{Work done during an internship at ByteDance}
 \\
University at Buffalo
\And
Huayi Zhang \\
ByteDance
\And
Peng Lin \\
ByteDance
\AND
Jinjun Xiong \\
University at Buffalo
\And
Wei Xu \\
ByteDance
}

\begin{document}

\ifcolmsubmission
\linenumbers
\fi

\maketitle

\begin{abstract}

We present a novel framework that improves the reliability of LLM judges by selectively augmenting LLM with auxiliary evaluation dimensions. Existing LLM judges often miss crucial evaluation dimensions because they fail to recognize the implicit standards underlying human assessments. To address this challenge, we propose the Auto-Prompt Ensemble (APE), an adaptive framework that automatically learns evaluation dimensions from its failure cases. APE incorporates a confidence-based ensemble mechanism to decide when to adopt the judgments from additional evaluation dimensions through a novel confidence estimation approach called \textit{Collective Confidence}. Extensive experiments demonstrate that APE improves the reliability of LLM Judge across diverse standard benchmarks. For instance, APE enhances GPT-4o’s agreement rate on \textsc{Reward Bench} from 87.2\% to 90.5\% in the zero-shot setting. Overall, APE provides a principled approach for LLM Judge to leverage test-time computation, and bridge the evaluation gap between human and LLM judges.

\end{abstract}
\section{Introduction}
Recent advances in large language models (LLMs) have enabled their use as evaluators—often referred to as "LLM Judges"\citep{zheng2023judgingllmasajudgemtbenchchatbot, kocmi2023largelanguagemodelsstateoftheart}. While LLMs have shown remarkable capabilities in complex domains such as mathematics and coding, their performance on tasks like text-quality evaluation, though ostensibly simpler, reveals a persistent gap in certain cases between model and human judgments. This discrepancy is particularly salient in domains where subjective, multi-dimensional criteria—such as coherence, humor, or stylistic appropriateness—play a central role. Previous efforts to improve LLM Judges have largely focused on supervised fine-tuning\citep{wang2024pandalm, zhu2025judgelmfinetunedlargelanguage, park2024offsetbias, ke2024critiquellminformativecritiquegeneration} or prompt engineering with powerful closed-source models~\citep{hu-etal-2024-llm-criteria, liu2024hdeval}, but these approaches rarely interrogate a deeper question: \textit{What limits the evaluation capabilities of LLM Judges?}

We argue that this limitation arises from two fundamental sources. First, LLMs must correctly identify the evaluation dimensions relevant to the task at hand—such as informativeness, fluency, or factuality—before any meaningful judgment can be made. This step, often implicitly assumed, is surprisingly brittle: poor prompt design or under-specified instructions can lead the model to overlook essential criteria~\citep{wang2023chatgptgoodnlgevaluator}. Crucially, many evaluation failures are not due to an inability to assess a property per se, but rather to a failure to recognize that the property is relevant in context. Second, even when the correct dimensions are identified, the model must make accurate comparative judgments along those axes. This may be constrained by its linguistic understanding, inductive biases, or training data artifacts~\citep{wang2023largelanguagemodelsfair}. While both components are important, we hypothesize that the primary bottleneck increasingly lies in the first: recognizing what should be evaluated. As models grow more capable, their failures are less often due to judgment inaccuracies and more often due to misalignment between the model’s inferred evaluation criteria and those intended by human judges.

To bridge the gap between human and LLM judges, we propose the \textbf{Auto-prompt Ensemble (APE)}—a novel evaluation framework that enhances LLM-based evaluation through automatic prompt augmentation and confidence-aware ensemble decision-making. APE begins with an \textbf{automated evaluation dimension generation} step, aimed at identifying and addressing gaps in the model’s understanding of the evaluation criteria. It first isolates failure cases—instances where the LLM judge’s initial output diverges from human annotations—thereby pinpointing situations where key dimensions may have been overlooked or misinterpreted. To resolve these discrepancies, APE employs the LLM itself to generate new evaluation dimensions, along with corresponding scoring rubrics, tailored to the failure context. For example, if the model fails to account for tone, humor, or factual consistency, the generation step can explicitly surface these elements as salient evaluation dimensions.

While these auxiliary dimensions enrich the LLM judge with task-adaptive evaluation criteria, a central challenge remains: how to coherently aggregate these diverse signals into a reliable final judgment. To this end, APE introduces a confidence-based ensemble mechanism that determines when to trust the collective input from a “jury” of evaluation dimensions. At the core of this mechanism is our novel \textbf{Collective Confidence} metric, which quantifies the agreement among individual judgments. Acting as a proxy for the ensemble’s overall reliability, collective confidence ensures that a final decision is made only when consensus is sufficiently strong, thereby improving both the accuracy and the trustworthiness of the evaluation process.

We conducted comprehensive experiments on established LLM judge benchmarks to assess the effectiveness of APE. We first applied APE to a 500-sample subset of the \textsc{Skywork Preference}\citep{liu2024skywork} dataset, where it dynamically generated 16 evaluation dimensions. This adaptation improved GPT-4o’s agreement with human annotations on the \textsc{Skywork Preference} test set from 83.6\% to 86.2\%, surpassing the majority-vote baseline of 84.5\%. We then transferred these dimensions to the \textsc{Reward Bench}\citep{lambert2024rewardbench}, resulting in a further improvement from 87.2\% to 90.5\%. Notably, APE outperformed the 84.5\% majority-vote baseline even in a zero-shot setting, demonstrating both strong generalization and computational efficiency. Overall, APE offers a principled framework for enhancing LLM judges through test-time adaptation, narrowing the gap between human and model evaluation standards while maintaining practical scalability.

\section{Auto-Prompt Ensemble}

APE is designed to address two core challenges:
\begin{itemize}
    \item \textbf{Evaluation Gaps between LLM Judges and Human Annotators.} LLM-based evaluations frequently miss critical dimensions valued by humans, causing significant misalignment with human annotations. To tackle this, we propose an automated framework that identifies missing evaluation dimensions by analyzing failure cases—situations where initial LLM judgments deviate from human annotations.
    
    \item \textbf{Selective Evaluation Dimension Ensemble.} Blindly incorporating newly discovered evaluation dimensions risks overriding accurate initial judgments. To prevent this, we propose a confidence-based ensemble strategy that selectively integrates additional evaluation dimensions -- overriding initial judgments only when jury consensus across multiple dimensions surpasses a calibrated confidence threshold.
\end{itemize}

Below, we provide an in-depth description of each component of our method. First, we detail how new evaluation dimensions are automatically generated to address the model’s specific failings (\S\ref{sec:auto-prompt}). Next, we explain how the auxiliary dimensions can be integrated via a collective confidence ensemble to better align with human judges (\S\ref{sec:collective-confidence}).

\subsection{Automatic Evaluation Dimension Generation}
\label{sec:auto-prompt}
\begin{figure}[t]
  \centering
  \includegraphics[width=\linewidth]{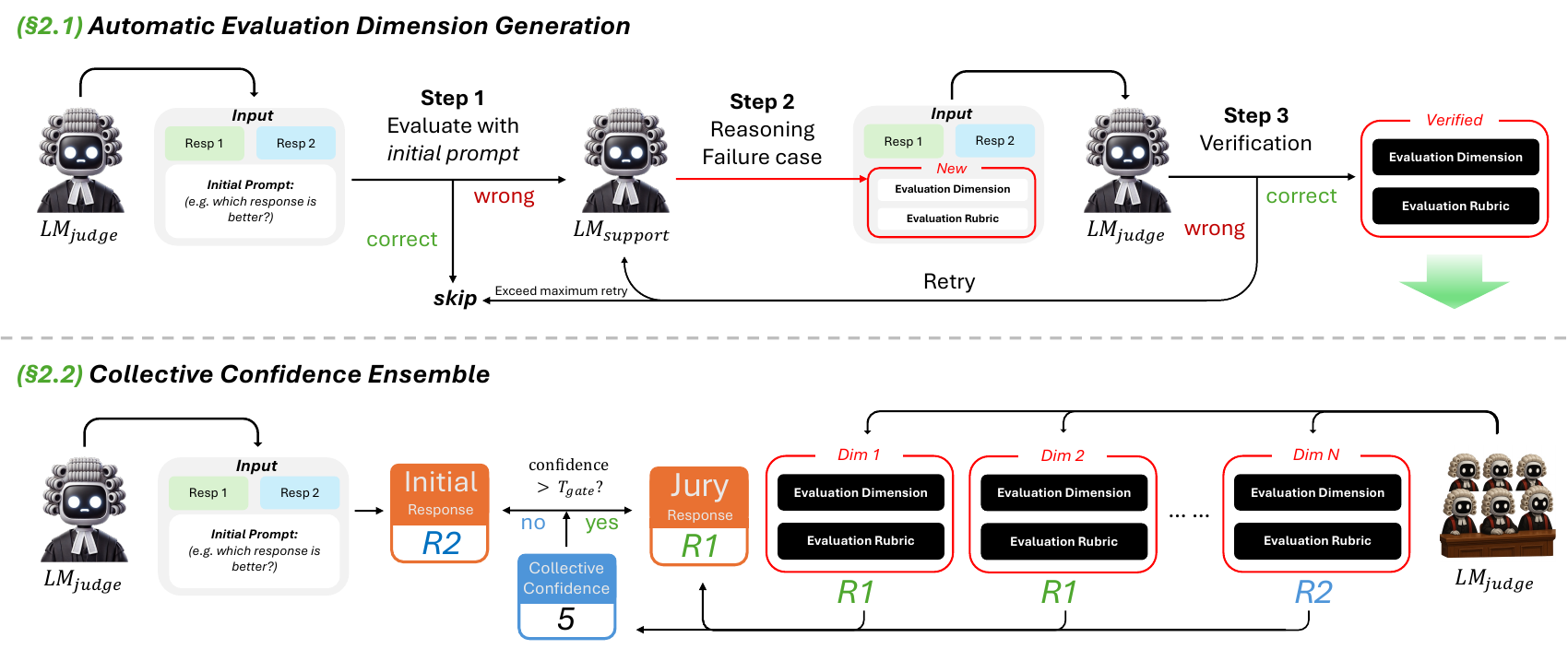}
  \caption{Overview of the APE framework. In the top pipeline, evaluation dimensions are automatically discovered by identifying failure cases and proposing targeted rubrics to correct them. In the bottom pipeline, a confidence-based ensemble aggregates judgments across verified dimensions, overriding the initial decision only when the collective confidence is sufficiently high.}


  \label{fig:pipeline}
\end{figure}

As illustrated in Fig.~\ref{fig:pipeline}, our approach proceeds in three steps to automatically generate evaluation dimensions. First, we detect failure cases where the Judge’s verdict conflicts with human annotations. Next, we generate and validate new evaluation dimensions that target these mistakes. Finally, we filter and select only those dimensions that consistently enhance alignment with human judgments, thereby refining our overall evaluation framework.

\paragraph{Collecting Failure Cases}
Given a training set \(D_{\text{train}} = \{(x_i, y_i)\}\), where \(x_i\) is the input (with two candidate responses) and \(y_i\) is the human-annotated ground-truth label, we prompt the Judge LLM, denoted \(\mathcal{LM}_{\text{judge}}\), to decide which response is superior for each instance \((x_i, y_i)\in D_{\text{train}}\). We collect failure cases where \(\mathcal{LM}_{\text{judge}}\)’s prediction is different from \(y_i\) and include them in a subset \(D_{\text{fail}} \subseteq D_{\text{train}}\) for further inspection.

\begin{figure}[t]
  \centering
  \includegraphics[width=\linewidth]{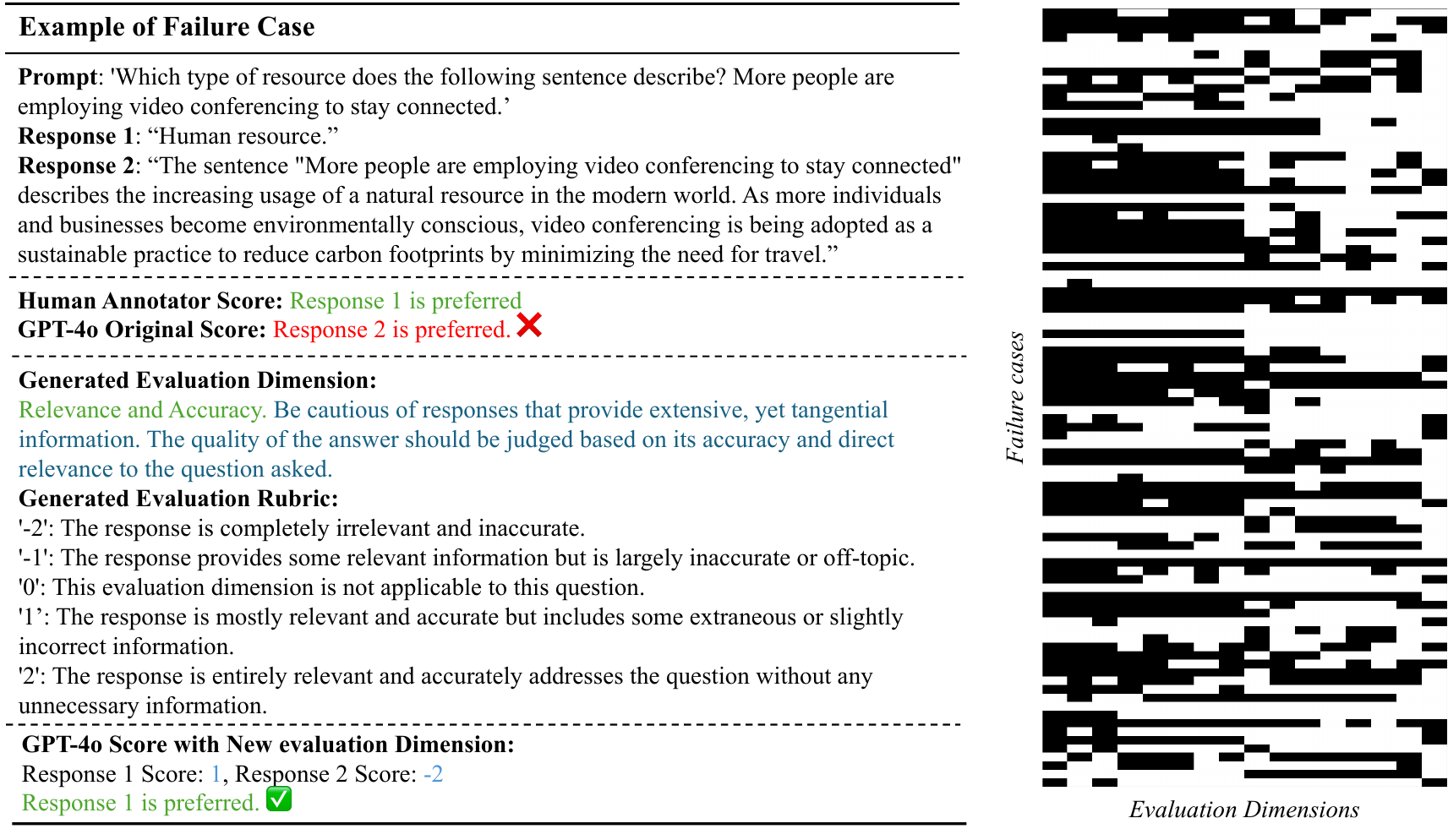}
  \caption{An example of a failure case from the \textsc{Skywork Reward Preference} dataset, where GPT-4o incorrectly prefers a suboptimal response. On the right, each cell indicates whether a newly generated evaluation dimension addresses the corresponding failure case: black denotes success, while white denotes failure. }
  \label{fig:failure_case}
\end{figure}

\paragraph{Generating and Validating Evaluation Dimensions}
For each failure case \((x_i, y_i) \in D_{\text{fail}}\), we invoke a supporting model \(\mathcal{LM}_{\text{support}}\) to diagnose the potential cause of error. It proposes a candidate evaluation dimension \(\delta_i\) along with a concise rubric describing how it should be applied. We incorporate \(\delta_i\) into \(\mathcal{LM}_{\text{judge}}\)'s prompt and re-evaluate the same instance \((x_i, y_i)\). If the revised prediction aligns with the human label \(y_i\), we consider \(\delta_i\) verified and add it to the candidate evaluation dimension set \(\Delta_{\text{verified}}\). Otherwise, \(\mathcal{LM}_{\text{support}}\) iteratively refines or replaces the proposed dimension, up to a fixed retry limit. If no verified dimension is found after exhausting the retry budget, the case is skipped. An example of a generated evaluation dimension is shown in Fig.~\ref{fig:failure_case}.

\paragraph{Dimensions Selection}
Once we have collected a set of verified dimensions \(\Delta_{\text{verified}} = \{\delta_1, \delta_2, \dots, \delta_m\}\), we reserve a subset of failure cases, \(D_{\text{val}} \subseteq D_{\text{fail}}\), as validation data. For each \(\delta_j \in \Delta_{\text{verified}}\) and every \((x_i, y_i) \in D_{\text{val}}\), we use evaluation dimension \(\delta_j\) to prompt \(\mathcal{LM}_{\text{judge}}\) to evaluate the failure case \(x_i\). Specifically, we define a binary indicator \(M_{ji}\) (with \(M_{ji}=1\) if the response generated under the prompt \(\delta_j\) aligns with the ground-truth label \(y_i\), and \(M_{ji}=0\) otherwise). The coverage rate for \(\delta_j\) is then computed as
\[
r_j = \frac{1}{|D_{\text{val}}|}\sum_{(x_i,y_i)\in D_{\text{val}}} M_{ji}.
\]
By ranking all dimensions according to their coverage rates \(r_j\), we select the top \(K\) dimensions—denoted \(\Delta^*\)—as our final evaluation dimensions.

\subsection{Collective Confidence Ensemble}
\label{sec:collective-confidence}
While the newly generated evaluation dimensions address recurring errors, their indiscriminate application may lead to overcorrection, potentially overriding correct initial judgments. To mitigate this risk, we propose a \textbf{Collective Confidence} mechanism that assesses the reliability of these additional dimensions and determines when to override the initial verdict.

\paragraph{Collective Confidence}
In existing confidence-estimation methods, either \textit{predictive probability} (the probability of the generated label or response~\citep{wang2022selfconsist_sc}) or \textit{verbal confidence} (prompting the model to provide a confidence score~\citep{lin2022teachingmodelsexpressuncertainty_verbal}) is used. However, as shown in Fig.~\ref{fig:ece}, both approaches can lead to overly concentrated confidence estimates, often clustering near the upper range (e.g., 0.8--1.0). Moreover, fine-tuning a strict threshold on such skewed distributions can drastically affect performance, especially in zero-shot scenarios. Chain-of-Thought reasoning can further inflate the final label’s probability by locking into a single reasoning path.

To address these challenges, we propose the \textbf{Collective Confidence} method, which treats each evaluation dimension as an independent juror. Suppose we have \(N\) evaluation dimensions, denoted as \(\Delta^* = \{\delta_i\}_{i=1}^N\). We consider a binary win/lose evaluation setting, where for a given input \(x\) and two candidate responses \(R_1\) and \(R_2\), the task is to determine which response is better. Each evaluation dimension (e.g., factuality, consistency, style) casts a vote by expressing its preference. Specifically, we define \(v_i\) for each dimension such that \(v_i = +1\) if \(\delta_i\) prefers \(R_1\) and \(v_i = -1\) if \(\delta_i\) prefers \(R_2\). We then calculate the aggregated jury confidence as follows:
\[
c_{\text{jury}} = \left|\sum_{i=1}^N v_i\right|.
\]
This absolute sum measures the degree of consensus among all dimensions. A larger \(c_{\text{jury}}\) indicates a stronger agreement. Finally, we map \(c_{\text{jury}}\) onto a calibrated scale \(\widetilde{c} \in [0.5, 1.0]\), where values closer to 1.0 signify strong agreement and values near 0.5 indicate a random guess. By relying solely on the consensus among multiple independent jurors, this approach minimizes the risk of overriding correct initial answers due to bias or overemphasis in any single evaluation dimension.

In Table~\ref{tab:ece}, we compare the proposed collective confidence approach with existing approaches, our method consistently outperforms both predictive probability and verbalized confidence on AUROC~\citep{fawcett2006introduction_auroc}, AUPRC~\citep{davis2006relationship_auprc}, and expected calibration error~\citep[ECE; ][]{guo2017calibrationmodernneuralnetworks_ece} on \textsc{Reward Bench}~\citep{lambert2024rewardbench} across all subsets. As shown in Fig.\ref{fig:ece}, the reliability plots indicate that collective confidence yields more calibrated estimates, with predicted confidence levels aligning closely with actual correctness rates—thereby serving as a robust signal of output reliability.

\paragraph{}
\begin{table}[t]
    \centering

\caption{Performance of confidence estimate approaches.}
\label{tab:ece}
\vspace{-0.3cm}
\renewcommand{\arraystretch}{1.2} 
\resizebox{\linewidth}{!}{
\begin{tabular}{cclcccccccccccc}
\toprule
\multicolumn{1}{l}{} & \multicolumn{1}{l}{}  & \multicolumn{1}{c}{\multirow{2}{*}{\textbf{Method}}} & \multicolumn{4}{c}{\textbf{Reward Bench}} & \multicolumn{4}{c}{\textbf{Donotanswer}} & \multicolumn{4}{c}{\textbf{LLMBar}}\\
\cmidrule(lr){4-7}  \cmidrule(lr){8-11} \cmidrule(lr){12-15}
\multicolumn{1}{l}{} & \multicolumn{1}{l}{}  && Acc.  & AUROC & AUPRC & ECE$\downarrow$ & Acc.  & AUROC & AUPRC & ECE$\downarrow$ & Acc. & AUROC & AUPRC & ECE$\downarrow$ \\
\midrule
\multirow{6}{*}{\textit{GPT-4o}} & \multirow{2}{*}{w/o CoT} & Predictive Probability & \textbf{0.852} & 0.927 & 0.925 & 0.290  & 0.634 & 0.708 & 0.691 & 0.297 & 0.685 & 0.728 & 0.712 & 0.376 \\
&  & Verbalized Confidence  & 0.880 & 0.917 & 0.890 & 0.235 & 0.681  & 0.734 & 0.688 & 0.354 & 0.780 & 0.797 & 0.754 & 0.137 \\
\cmidrule(lr){2-15} 
& \multirow{3}{*}{w/ CoT}  & Predictive Probability & 0.871 & 0.910 & 0.890 & 0.261 & 0.711 & 0.763 & 0.744 & 0.288 & 0.770 & 0.802 & 0.768 & 0.363 \\
&  & Verbalized Confidence  & 0.879 & 0.922 & 0.896 & 0.317 & 0.681  & 0.717 & 0.670 & 0.488 & 0.783 & 0.838 & 0.803 & 0.164 \\
&  & Collective Confidence (Ours) & 0.863 (\textbf{0.905}) & \textbf{0.934} & \textbf{0.927} & \textbf{0.041} & \textbf{0.763}  & \textbf{0.857} & \textbf{0.854} & \textbf{0.124} & \textbf{0.876}  & \textbf{0.931} & \textbf{0.919} & \textbf{0.098}\\
\bottomrule
\end{tabular}
}

    \vspace{0.2cm}
  \centering
  \begin{subtable}{0.3\linewidth}
    \centering
    \includegraphics[width=\linewidth]{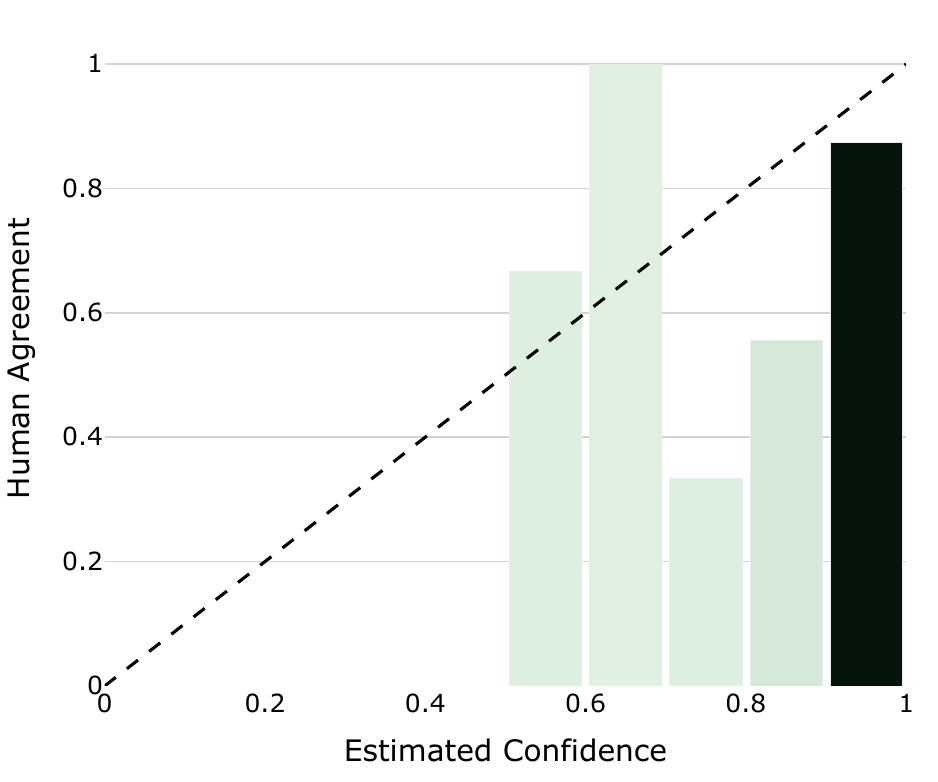}
    \caption{Predictive Probability}
    \label{fig:third}
  \end{subtable}
  \hfill
  \begin{subtable}{0.3\linewidth}
    \centering
    \includegraphics[width=\linewidth]{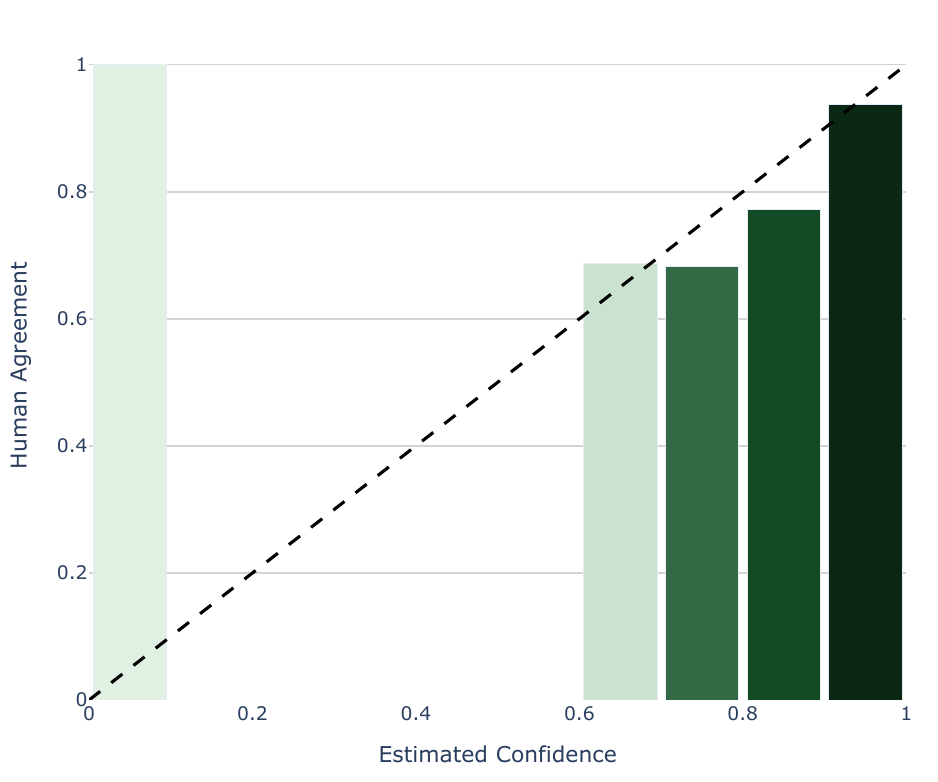}
    \caption{Verbalized Confidence}
    \label{fig:second}
  \end{subtable}
  \hfill
  \begin{subfigure}{0.3\linewidth}
    \centering
    \includegraphics[width=\linewidth]{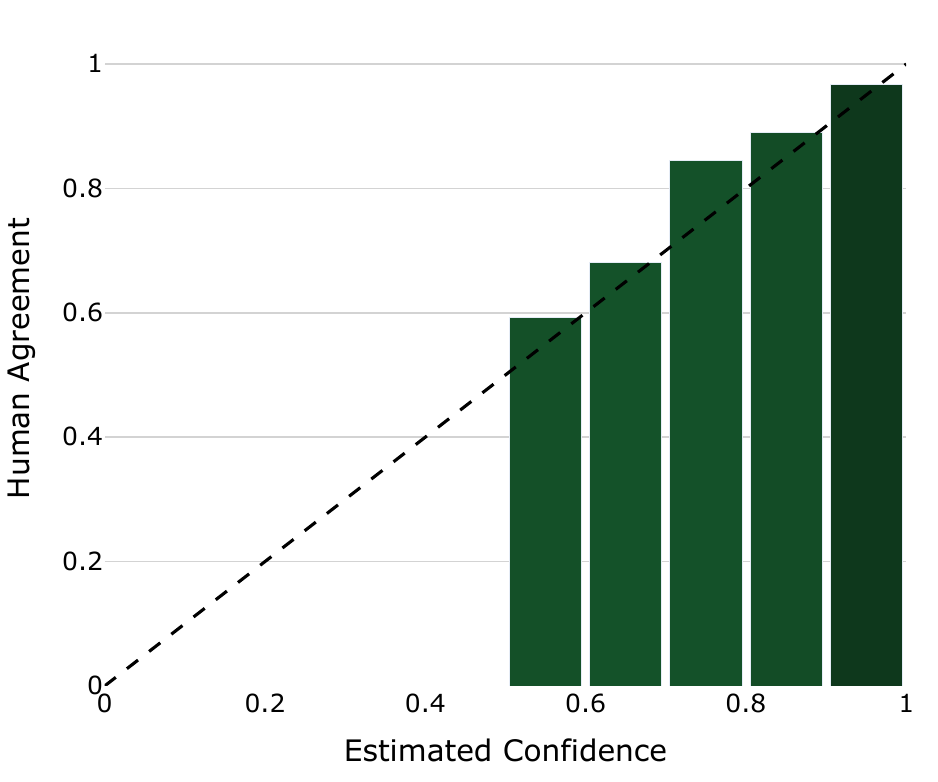}
    \caption{Collective Confidence}
    \label{fig:first}
  \end{subfigure}
  \vspace{-0.5cm}
  \captionof{figure}{Reliability plot for confidence estimation methods. Using GPT-4o with CoT as Judge on \textsc{Reward Bench}. A deep color indicates a higher percentage. The dashed diagonal line represents perfect calibration, where estimated confidence matches actual agreement.}
  \label{fig:ece}

    \label{tab:combined_ece}
\end{table}

\paragraph{Ensemble Decision Strategy}
With a collective confidence measure in place, we must decide when to replace the initial verdict \(\widehat{R}\). Let the jury’s majority preference be \(\widehat{R}_{\text{jury}} = R_1\) if \(\sum_{i=1}^N v_i > 0\), and \(R_2\) otherwise. We then combine both initial and jury decisions via:
\[
R_{\text{final}} =
\begin{cases}
\widehat{R}_{\text{jury}}, & \text{if } c_{\text{jury}} > T_{\text{gate}},\\
\widehat{R}, & \text{otherwise},
\end{cases}
\]
where \(T_{\text{gate}}\) is set using a small calibration set \(D_{\text{cal}}\). If consensus among dimensions does not surpass this threshold, we retain the initial decision, thus preserving correct judgments that do not need extra intervention. We find that even modest calibration data is sufficient to select a robust threshold, rendering our ensemble approach versatile across various tasks.

\section{Experiments}

\subsection{Experimental Setup}

\paragraph{Benchmarks} We evaluate the performance of our LLM Judge system on four standard benchmarks. First, we utilize the \textsc{Reward Bench} dataset~\citep{lambert2024rewardbench}, which covers a diverse range of tasks such as chatting, challenging conversations, safety, and reasoning (code and math). The dataset assesses models by comparing the scores assigned to preferred versus rejected responses. Notably, in several of its subsets, the baseline performance of advanced models like GPT-4o is nearly saturated (exceeding 95\%). Therefore, we focus our experiments on the subsets that have not yet reached saturation: \textit{LLMBar}~\citep{zeng2024evaluatinglargelanguagemodels_llmbar}, \textit{Donotanswer}~\citep{wang2024chinese_donotanswer}, \textit{AlpacaEval}~\citep{alpaca_eval}, \textit{MT-Bench}~\citep{zheng2023judgingllmasajudgemtbenchchatbot}, \textit{XSTest}~\citep{röttger2024xstesttestsuiteidentifying}, and \textit{Refusals}
\footnote{To ensure the evaluation centers on a model's ability to judge response quality rather than its domain-specific capabilities, we remove math- and code-related questions from the benchmarks. These tasks require the model itself to possess strong problem-solving abilities in coding and mathematics, which deviates from the core goal of evaluating judgment performance. Furthermore, models like GPT-4o have already achieved saturated accuracy on such questions in Reward Bench.}.
Second, we employ the \textsc{Skywork Reward Preference} dataset~\citep{liu2024skywork}, which comprises 80K preference pairs subsampled from multiple public datasets, including \textit{HelpSteer2}~\citep{wang2024helpsteer2opensourcedatasettraining}, \textit{OffsetBias}~\citep{park2024offsetbias}, and \textit{WildGuard}~\citep{wildguard2024}. From this dataset, we randomly sample 500 training samples, 200 calibration samples, and 1000 test samples.

\paragraph{Models and baselines} We conduct experiments using three different models: GPT-4o~\citep{openai2024gpt4ocard}, as well as the Qwen-2.5-7B-Intruct and Qwen-2.5-72B-Intruct ~\citep{qwen2025qwen25technicalreport}. We use the temperature of 0.6 for all experiments. These models allow us to assess the effectiveness of our method across a range of model scales and architectures. We compare our proposed method against several baselines. The first baseline, \textit{Vanilla}, employs a general LLM Judge with a standard prompt. The second baseline, \textit{Majority Vote}~\citep{wang2022selfconsist_sc}, aggregates judgments across multiple criteria. We provide the full list of prompts used in our experiments in Appendix~\ref{app:prompt}.

\paragraph{Metrics} 
We report \textit{agreement rate} as our primary evaluation metric, defined as the percentage of model predictions that match the human-annotated ground-truth preferences. This metric reflects how well an LLM-based judge aligns with human judgment and is commonly used in prior work on preference modeling and response evaluation.

\subsection{Automatic Evaluation Dimension Generation}
We performed automatic evaluation dimension generation on a 500-sample subset of the \textsc{Skywork Reward Preference} dataset. Using the GPT-4o as Judge LLM, we identified 92 failure cases—instances where the model's judgment disagreed with the human-labeled ground truth—corresponding to an initial agreement rate of 81.6\%. For each failure case, we employed GPT-4o as a reasoning model to propose corrective evaluation dimensions, allowing up to 10 attempts per case. This process produced 88 candidate dimensions in total.
To ensure the quality of the generated evaluation dimensions, we applied a semantic-based filtering step based on scoring separation. Specifically, for each candidate dimension, we prompted the Judge LLM to independently score both responses in each failure case according to the provided rubric. We then computed the absolute difference between the two scores. Dimensions for which the score difference did not exceed a threshold of 2 were discarded, as such dimensions often reflect vague or ambiguous criteria that fail to meaningfully distinguish between responses. This filtering process reduced the set from 88 to 40 verified evaluation dimensions.
We then measured the effectiveness of each dimension by prompting GPT-4o on the original 92 failure cases. Finally, we selected the top 16 dimensions with the highest coverage rates, achieving a post-selection coverage rate of 84.8\% on the failure set $D_{\text{fail}}$. Threshold calibration was then performed on the 200 calibration samples, and $T_{\text{gate}} = 4$ was selected for use in subsequent experiments.

\subsection{Main Results}

\paragraph{Skywork Reward Preference} 
As shown in Table~\ref{tab:combined_ece}, APE@16 achieves the highest overall agreement rate of 86.2\%, outperforming both the Vanilla GPT-4o Judge (83.0\%) and the Majority Vote baseline (84.5\%). The Majority Vote method, which ensembles judgments across multiple static criteria, serves as a strong baseline for evaluating the effectiveness of automatically discovered evaluation dimensions. APE notably improves performance on the \textit{OffsetBias} subset (91.0\% vs.\ 81.5\%), suggesting its ability to capture subtle, bias-sensitive evaluation aspects that are often missed by fixed criteria. Similar trends hold for \textit{HelpSteer2} and \textit{WildGuard}, demonstrating that auto-prompted dimensions transfer well across heterogeneous preference sources.
\begin{table}[t]
\centering
\caption{Results on \textsc{Skywork Reward Preference}.}
\vspace{-0.2cm}
\setlength{\tabcolsep}{9pt}
\resizebox{\textwidth}{!}{%
\begin{tabular}{lcc cccc}
\toprule
\multirow{2}{*}{\textbf{Method}} & \multirow{2}{*}{\textbf{Auto-prompt}} & \multirow{2}{*}{\textbf{Collective Confidence}} & \multicolumn{4}{c}{\textbf{Agreement Rate}} \\
\cmidrule(lr){4-7}
& & & \textbf{HelpSteer2} & \textbf{OffsetBias} & \textbf{Wildguard} & \textbf{All} \\
\midrule
\textbf{GPT-4o} &&&&&& \\
Vanilla         & - & - & 71.7 & 81.5 & 92.7 & 83.0 \\
Majority Vote @ 16  & - & - & \textbf{75.9} & 81.5 & \textbf{94.2} &  84.5 \\
\rowcolor{green!10}   & \checkmark & -    & 69.6 & 86.6 & 92.6 & 85.1 \\
\rowcolor{green!10} \multirow{-2}{*}{APE @ 16 (Ours)}    & \checkmark & \checkmark & 72.2 & \textbf{91.0} & 90.0 & \textbf{86.2} \\
\bottomrule
\label{tab:main}
\end{tabular}
}

\centering
\caption{Results on \textsc{Reward Bench}.}
\vspace{-0.2cm}
\setlength{\tabcolsep}{5pt}
\resizebox{\textwidth}{!}{%
\begin{tabular}{lcc ccccccc}
\toprule
\multirow{2}{*}{\textbf{Method}} & \multirow{2}{*}{\textbf{Auto-prompt}} & \multirow{2}{*}{\textbf{Collective Confidence}} & \multicolumn{7}{c}{\textbf{Agreement Rate}} \\
\cmidrule(lr){4-10}
& & & \textbf{LLMBar} & \textbf{Donotanswer} & \textbf{AlpacaEval} & \textbf{MT-Bench} & \textbf{XSTest} & \textbf{Refusals} & \textbf{All}\\
\midrule
\textbf{GPT-4o} &&&&&&&&& \\
Vanilla         & - & - & 71.6 & 71.2 & \textbf{96.6} & 95.2 & 94.8 & \textbf{97.0} & 87.2 \\
Majority Vote @ 16   & - & - & 72.6 & 72.6 & 96.6 & 95.2 & 94.3 & 97.0 & 87.4 \\
Monolithic Prompt   & \checkmark  & - & 82.6 & 79.4 & 80.0 & 92.4 & 94.1 & 94.0 & 86.9 \\
In-context Learning  & \checkmark  & -  & 71.8 & 69.9 & 85.9 & 79.0 & 89.4 & 93.0 & 82.0 \\

\rowcolor{green!10}             & \checkmark & -         & \textbf{87.8} & \textbf{77.0} & 77.2 & 90.5 & 95.3  & 86.0 & 86.8 \\
\rowcolor{green!10} \multirow{-2}{*}{APE @ 16 (Ours)}  &  \checkmark & \checkmark  & 85.2 & 74.8 & 92.8 & \textbf{95.2} & \textbf{96.0} & 95.0 & \textbf{90.5} \\
\midrule
\textbf{Qwen-2.5-7B-Instruct} &&&&&&&&& \\
Vanilla         & - & - & 61.3 & 71.9 & \textbf{90.3} & \textbf{90.5} & 79.2 & 97.5 & 79.0 \\
\rowcolor{green!10} APE @ 16 (Ours) 
                & \checkmark & \checkmark  & \textbf{66.8} & \textbf{77.0} & 89.3 & 89.5 & \textbf{82.7} & \textbf{98.5} & \textbf{81.7} \\
\midrule
\textbf{Qwen-2.5-72B-Instruct} &&&&&&&&& \\
Vanilla         & - & - & 63.7 & 75.6 & 94.8 & 94.3 & 88.6 & \textbf{96.5} & 83.3 \\
\rowcolor{green!10} APE @ 16 (Ours) 
                & \checkmark & \checkmark  & \textbf{72.6} & \textbf{75.6} & \textbf{95.2} & \textbf{96.2} & \textbf{93.3} & 94.0 & \textbf{86.8} \\
\bottomrule
\label{tab:llmbar}
\end{tabular}
}
\end{table}



\paragraph{Reward Bench (Zero-shot)} 
As shown in Table~\ref{tab:combined_ece}, APE@16 also demonstrates consistent improvements across challenging subsets in a zero-shot setting. Here, all evaluation dimensions used by APE are generated from the Skywork Reward Preference dataset and directly transferred to Reward Bench without any task-specific tuning. Despite this cross-domain shift, APE achieves 92.8\% on \textit{AlpacaEval}, 96.0\% on \textit{XSTest}, and 95.0\% on \textit{Refusals}, resulting in an overall agreement of 90.5\%, outperforming both the Vanilla Judge (87.2\%) and Majority Vote (87.4\%). These gains indicate that APE effectively augments the evaluation process with context-sensitive criteria that generalize well across domains. Moreover, the improvements on high-performing subsets underscore the value of APE’s collective confidence strategy in avoiding unnecessary overrides, making it a robust and practical solution even when the base judge already performs strongly. Importantly, since no in-distribution data is required from the target benchmark, APE significantly improves test-time scalability by enabling judgment enhancement without dataset-specific adaptation.

\paragraph{Ablation Study} 
To disentangle the contributions of each component, we conduct an ablation study focusing on automatic prompt generation and collective confidence. When using only auto-prompting, APE@16 already outperforms both the Vanilla Judge and Majority Vote on key subsets like \textit{LLMBar} (87.8\%) and \textit{XSTest} (95.3\%), reaching an overall agreement of 86.8\%. This provides further evidence that automatically identified dimensions are more effective than fixed voting criteria. Adding the collective confidence mechanism further raises performance to 90.5\%, confirming that the ensemble strategy helps selectively apply dimensions and improves overall judgment reliability. 

\paragraph{Evaluation Dimension Transfer}
To assess the transferability of learned evaluation criteria across models, we apply the evaluation dimensions generated by GPT-4o on Skywork to different models, including Qwen-2.5-7B-Instruct and Qwen-2.5-72B-Instruct. As shown in Table~\ref{tab:combined_ece}, despite architectural and scale differences, both Qwen models benefit from the same auto-prompted dimensions without any retraining or adaptation. Specifically, for Qwen-2.5-7B-Instruct, APE@16 improves the overall agreement rate from 79.0\% (Vanilla) to 81.7\%, with substantial gains on subsets such as \textit{Donotanswer} (from 71.9\% to 77.0\%) and \textit{Refusals} (from 97.5\% to 98.5\%). Similarly, for Qwen-2.5-72B-Instruct, APE@16 improves the overall agreement rate from 83.3\% to 86.8\%, with strong improvements on \textit{AlpacaEval} (from 94.8\% to 95.2\%) and \textit{MT-Bench} (from 94.3\% to 96.2\%). These results demonstrate that evaluation dimensions are not only robust across datasets but also transferable across model sizes and architectures, highlighting their potential for reusable evaluation augmentation in practical deployment scenarios.

\begin{figure}[t]
  \centering
  \begin{subfigure}[t]{0.45\linewidth}
    \centering
    \includegraphics[width=\linewidth]{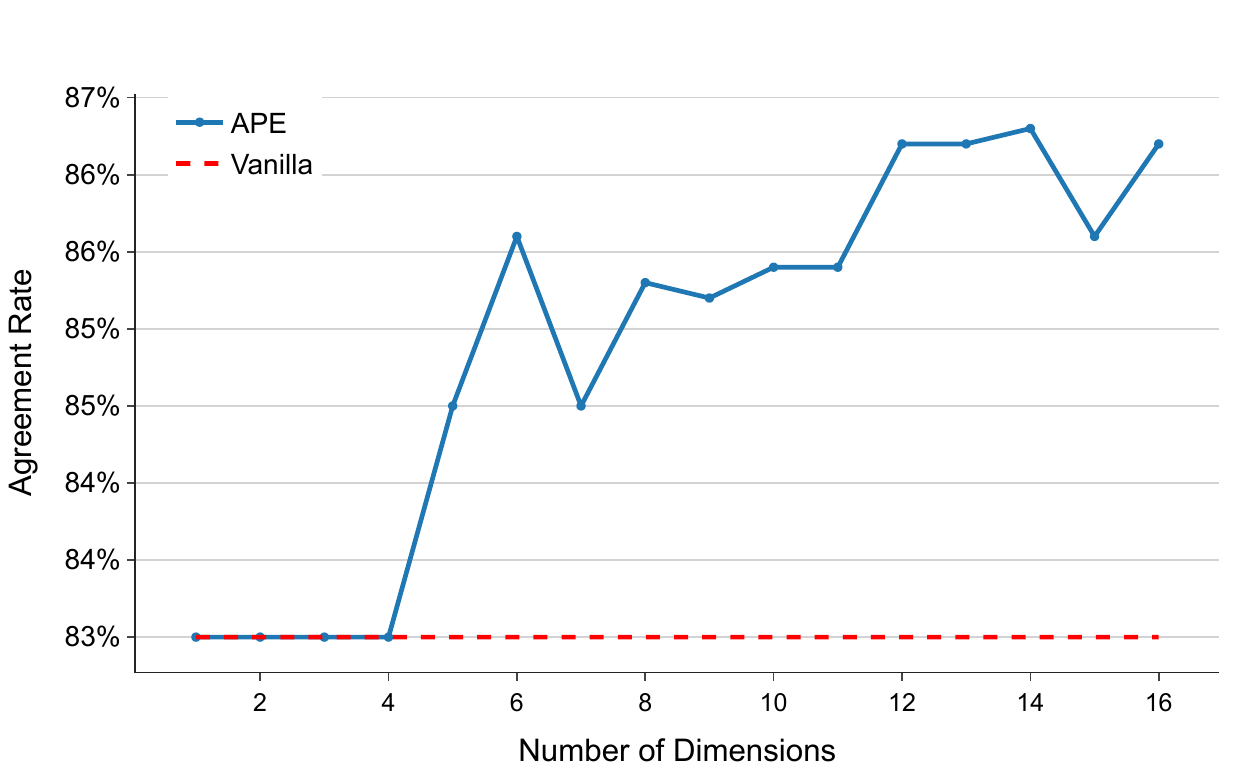}
    \caption{\textsc{Skywork Preference}}
    \label{fig:subfig1}
  \end{subfigure}
  \hfill
  \begin{subfigure}[t]{0.45\linewidth}
    \centering
    \includegraphics[width=\linewidth]{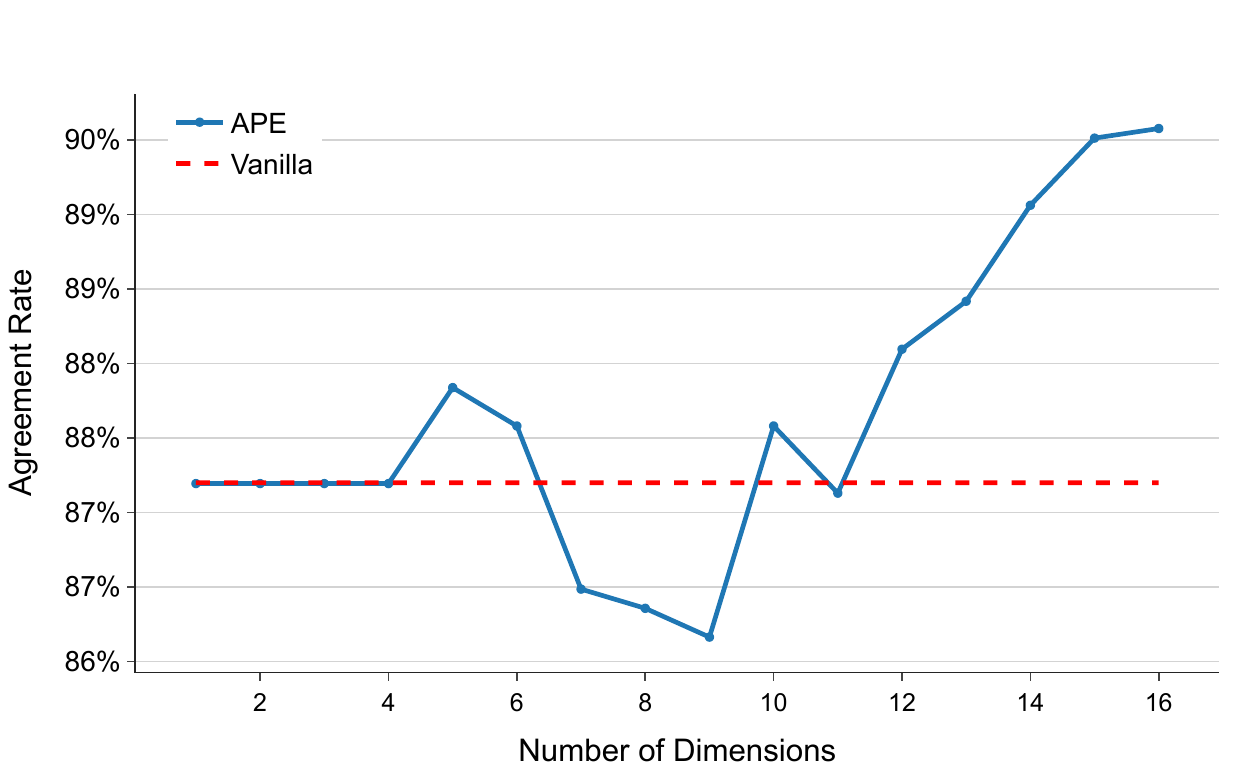}
    \caption{\textsc{Reward Bench}}
    \label{fig:subfig2}
  \end{subfigure}
  \caption{Impact of Number of Dimensions}
  \label{fig:num_dimensions}
\end{figure}

\paragraph{Impact of Number of Evaluation Dimensions} 
Figure~\ref{fig:num_dimensions} illustrates how the number of evaluation dimensions affects agreement rates on both the Skywork Preference and Reward Bench datasets. On Skywork (Figure~\ref{fig:num_dimensions}a), performance begins to improve noticeably once more than 4 dimensions are introduced, with a steady upward trend observed as additional dimensions are incorporated. The agreement rate peaks around 14--16 dimensions, indicating that the framework benefits from a richer and more diverse set of evaluation perspectives. In contrast, the zero-shot Reward Bench results (Figure~\ref{fig:num_dimensions}b) exhibit a different pattern. While early additions of dimensions do not lead to immediate improvements—and may even introduce mild degradation due to distribution shift—performance increases significantly after 10 dimensions, eventually surpassing the baseline. This suggests that although dimension transfer can initially be noisy in out-of-distribution settings, a sufficient number of robust and generalizable dimensions can ultimately yield substantial gains. Overall, these results highlight the importance of both quantity and quality in selecting evaluation dimensions for effective judgment enhancement.

\section{Related Works}

\paragraph{LLM Judge}
LLMs are increasingly employed as automatic evaluators—so-called “LLM Judges”~\cite{zheng2023judgingllmasajudgemtbenchchatbot}. However, despite their capabilities, these models often fall short in seemingly simpler tasks such as evaluating the quality of natural language text. This performance gap has sparked growing interest in enhancing LLMs' evaluation abilities. Prior work has primarily focused on supervised fine-tuning \citep{wang2024pandalm, zhu2025judgelmfinetunedlargelanguage, park2024offsetbias, ke2024critiquellminformativecritiquegeneration}, prompt engineering \citep{zheng2023judgingllmasajudgemtbenchchatbot}, or ensembling human-written criteria \citep{hu-etal-2024-llm-criteria}. Others, like \citet{liu2024hdeval}, leverage hierarchical prompts and inference pipelines to improve evaluation granularity. In contrast, APE introduces a novel approach: an automated pipeline that identifies and learns from failure cases to generate new evaluation dimensions, aiming to address LLMs’ persistent weaknesses in judgment tasks.


\paragraph{Automatic Prompt Engineering}
Automatic Prompt Engineering (AutoPE) aims to reduce the reliance on manually crafted prompts by autonomously generating and refining prompts to enhance the performance of LLMs. \cite{shin2020autopromptelicitingknowledgelanguage} proposed AutoPrompt, a technique that automatically generates prompts through gradient-guided search. \cite{zhou2022large_ape} introduced the AutoPE framework, which formulates instruction generation as a natural language synthesis problem addressed through black-box optimization, leveraging LLMs to propose and evaluate candidate instructions.
 Furthermore, \cite{guo2023learningplannaturallanguage_apo2} explored methodologies for learning to plan using natural language, emphasizing the role of automatic prompt engineering in enhancing models' capabilities to understand and generate complex plans.
Similarly, \cite{pryzant2023automaticpromptoptimizationgradient_apo} presented Automatic Prompt Optimization (APO), a method inspired by numerical gradient descent to automatically refine prompts from failure cases. Our method differs from APO in that, rather than repeatedly refining a single prompt, we adopt a confidence-based ensemble strategy that integrates multiple gradient-derived candidates. 


\paragraph{Confidence Estimation in LLMs}
Accurate confidence estimation is critical for trustworthy LLM-based evaluation, especially when model predictions are used in high-stakes or alignment-sensitive settings. Prior work has explored two main approaches: (1) \textit{predictive probability}, which relies on the model's output distribution over labels~\citep{wang2022selfconsist_sc, kadavath2022languagemodelsmostlyknow, jung2024trust}; and (2) \textit{verbalized confidence}, where the model is explicitly prompted to express its confidence in natural language~\citep{lin2022teachingmodelsexpressuncertainty_verbal}. While intuitive, these methods often produce overconfident or poorly calibrated estimates. Predictive probabilities, in particular, are known to be overconfident in large models~\citep{guo2017calibrationmodernneuralnetworks_ece}, and can be further distorted by reasoning strategies like Chain-of-Thought prompting~\citep{wei2023chain, turpin2023languagemodelsdontsay}. These limitations highlight the need for more robust, interpretable confidence estimation mechanisms—especially in comparative evaluation scenarios where nuanced preferences must be inferred across multiple dimensions.

\section{Conclusion}
We present Auto-Prompt Ensemble, a framework that improves LLM-based evaluation by addressing a key limitation: LLMs often miss essential evaluation dimensions that humans consider. Our method detects low-confidence cases, generates new task-specific evaluation prompts from real failure examples, and uses a selective ensemble to override initial judgments only when there is strong multi-dimensional agreement. Experiments across benchmarks show that APE significantly boosts alignment with human preferences, even in zero-shot and cross-model settings. These results underscore our core insight: the primary challenge in LLM judgment is not misapplying criteria, but failing to identify which criteria matter. APE closes this gap, offering a scalable and transferable way to build more accurate and trustworthy LLM evaluators.

\bibliography{colm2025_conference}
\bibliographystyle{colm2025_conference}

\newpage
\appendix
\section{Appendix}
\label{app:prompt}

\begin{figure*}[h]
  \centering
  \caption{Prompt used for LLM Judge inference.}
  \vspace{-0.1cm}
  \includegraphics[width=\linewidth]{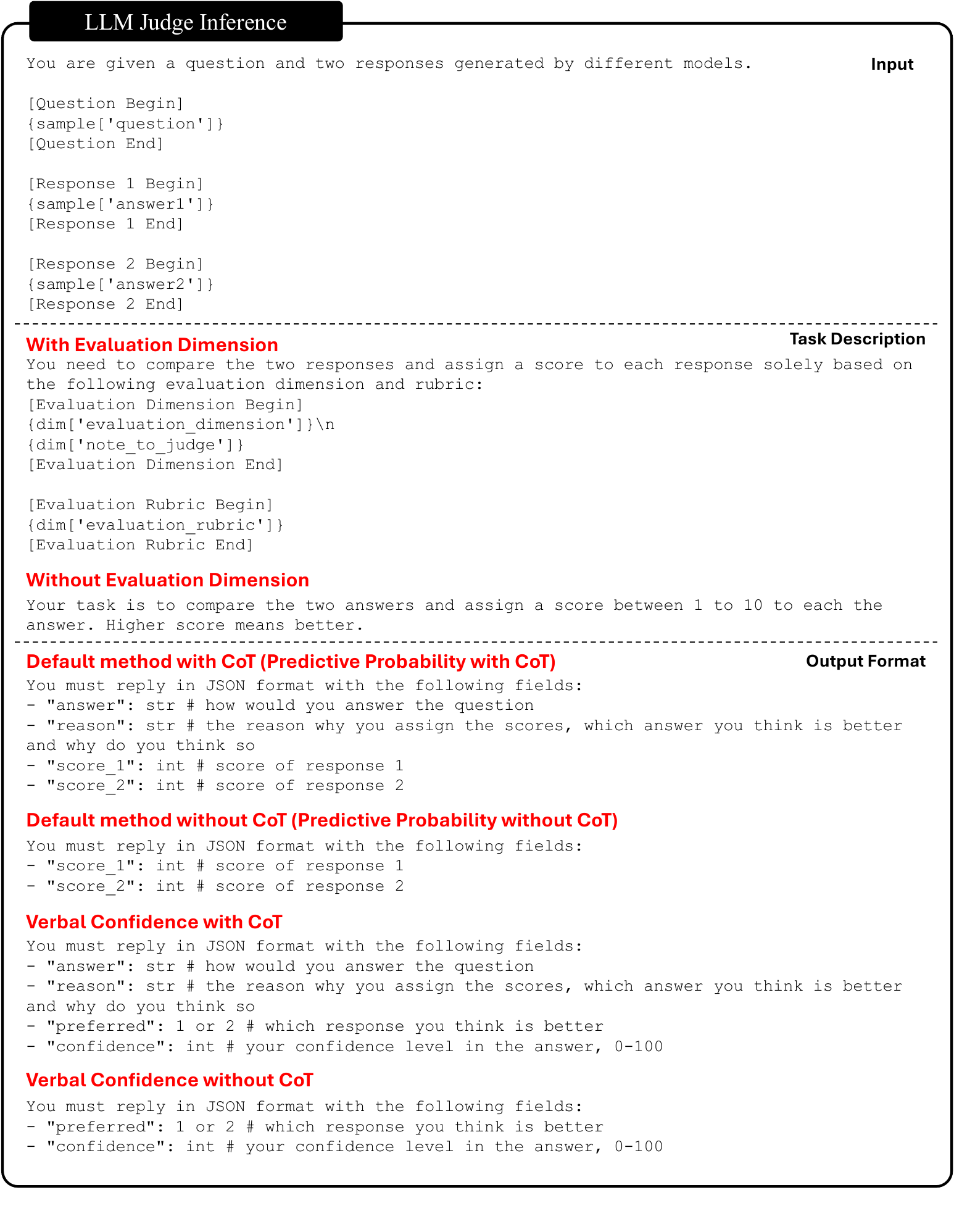}
  \label{tab:prompt_inference}
\end{figure*}

In our evaluation framework, the evaluation dimensions are generated using the prompt outlined in Table~\ref{tab:prompt_dimgen}, while the inference is conducted using the prompt detailed in Table~\ref{tab:prompt_inference}. Each prompt comprises an input section containing the question and corresponding model responses, a detailed task description, and an output format that typically requires a JSON structure with fields like an evaluation rubric, or the score of a response.

\begin{table*}[h]
  \centering
  \caption{Prompt used to generate evaluation dimensions.}
  \vspace{-0.2cm}
  \includegraphics[width=\linewidth]{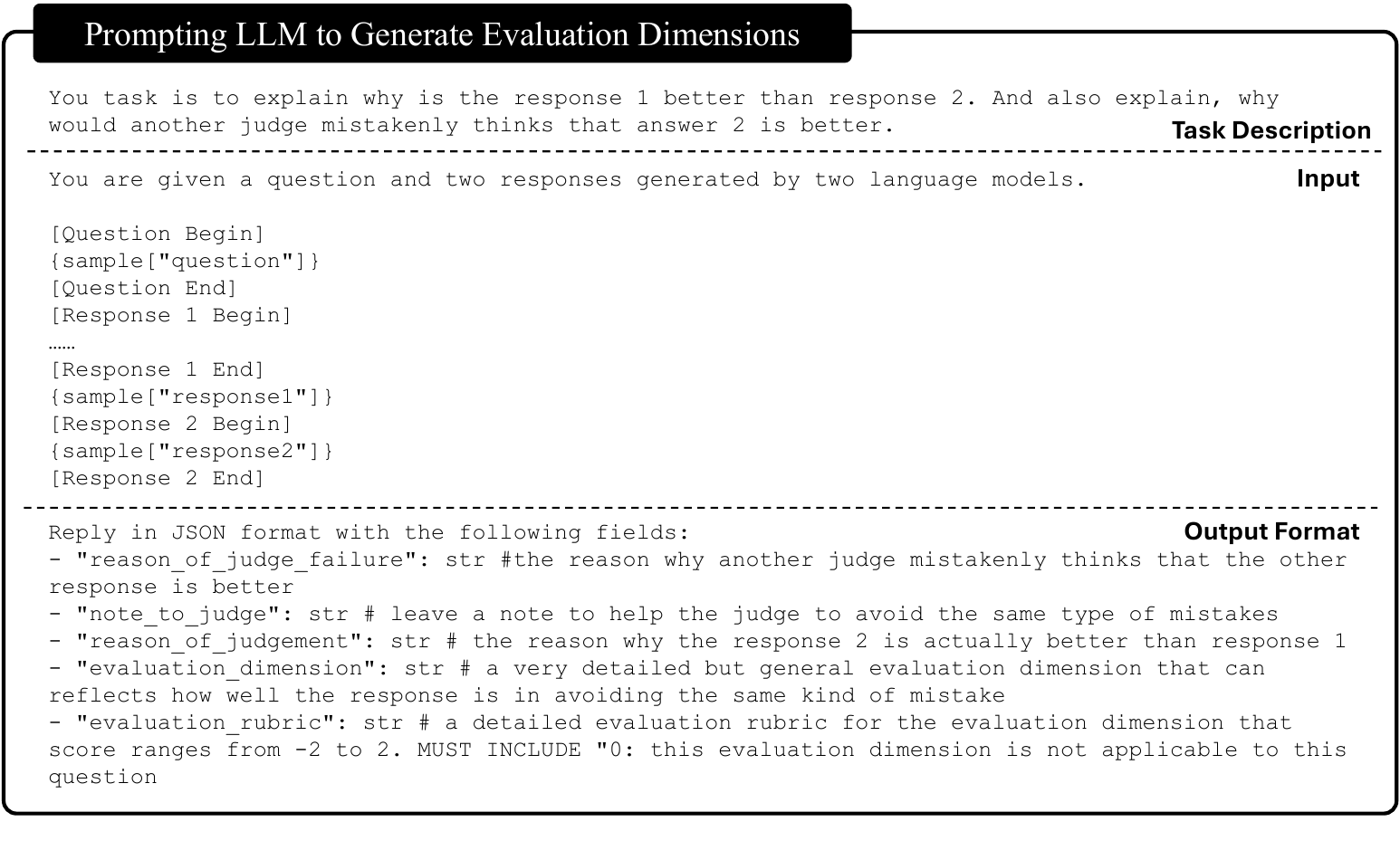}
  \label{tab:prompt_dimgen}
\end{table*}


\end{document}